# Evolutionary Algorithms for Reinforcement Learning

**David E. Moriarty**                                    MORIARTY@ISI.EDU
*University of Southern California, Information Sciences Institute*
*4676 Admiralty Way, Marina del Rey, CA 90292*

**Alan C. Schultz**                                    SCHULTZ@AIC.NRL.NAVY.MIL
*Navy Center for Applied Research in Artificial Intelligence*
*Naval Research Laboratory, Washington DC 20375-5337*

**John J. Grefenstette**                                    GREF@IB3.GMU.EDU
*Institute for Biosciences, Bioinformatics and Biotechnology*
*George Mason University, Manassas, VA 20110*

## Abstract

There are two distinct approaches to solving reinforcement learning problems, namely, searching in value function space and searching in policy space. Temporal difference methods and evolutionary algorithms are well-known examples of these approaches. Kaelbling, Littman and Moore recently provided an informative survey of temporal difference methods. This article focuses on the application of evolutionary algorithms to the reinforcement learning problem, emphasizing alternative policy representations, credit assignment methods, and problem-specific genetic operators. Strengths and weaknesses of the evolutionary approach to reinforcement learning are presented, along with a survey of representative applications.

## 1. Introduction

Kaelbling, Littman, and Moore (1996) and more recently Sutton and Barto (1998) provide informative surveys of the field of reinforcement learning (RL). They characterize two classes of methods for reinforcement learning: methods that search the space of value functions and methods that search the space of policies. The former class is exemplified by the temporal difference (TD) method and the latter by the evolutionary algorithm (EA) approach. Kaelbling et al. focus entirely on the first set of methods and they provide an excellent account of the state of the art in TD learning. This article is intended to round out the picture by addressing evolutionary methods for solving the reinforcement learning problem.

As Kaelbling et al. clearly illustrate, reinforcement learning presents a challenging array of difficulties in the process of scaling up to realistic tasks, including problems associated with very large state spaces, partially observable states, rarely occurring states, and non-stationary environments. At this point, which approach is best remains an open question, so it is sensible to pursue parallel lines of research on alternative methods. While it is beyond the scope of this article to address whether it is better in general to search value function space or policy space, we do hope to highlight some of the strengths of the evolutionary approach to the reinforcement learning problem. The reader is advised not to view this article as an EA *vs.* TD discussion. In some cases, the two methods provide complementary strengths, so hybrid approaches are advisable; in fact, our survey of implemented systems illustrates that many EA-based reinforcement learning systems include elements of TD-learning as well.

The next section spells out the reinforcement learning problem. In order to provide a specific anchor for the later discussion, Section 3 presents a particular TD method. Section 4 outlines the approach we call Evolutionary Algorithms for Reinforcement Learning (EARL), and provides a simple example of a particular EARL system. The following three sections focus on features that distinguish EAs for RL from EAs for general function optimization, including alternative policy representations, credit assignment methods, and RL-specific genetic operators. Sections 8 and 9 highlight some strengths and weaknesses of the EA approach. Section 10 briefly surveys some successful applications of EA systems on challenging RL tasks. The final section summarizes our presentation and points out directions for further research.

## 2. Reinforcement Learning

All reinforcement learning methods share the same goal: to solve *sequential decision tasks* through trial and error interactions with the environment (Barto, Sutton, & Watkins, 1990; Grefenstette, Ramsey, & Schultz, 1990). In a sequential decision task, an agent interacts with a dynamic system by selecting actions that affect state transitions to optimize some reward function. More formally, at any given time step $t$, an agent perceives its *state* $s_t$ and selects an *action* $a_t$. The system responds by giving the agent some (possibly zero) numerical *reward* $r(s_t)$ and changing into state $s_{t+1} = \delta(s_t, a_t)$. The state transition may be determined solely by the current state and the agent's action or may also involve stochastic processes.

The agent's goal is to learn a *policy*, $\pi : S \rightarrow A$, which maps states to actions. The *optimal policy*, $\pi^*$, can be defined in many ways, but is typically defined as the policy that produces the greatest cumulative reward over all states $s$:

$$\pi^* = \underset{\pi}{\operatorname{argmax}} \, V^\pi(s), (\forall s) \tag{1}$$

where $V^\pi(s)$ is the cumulative reward received from state $s$ using policy $\pi$. There are also many ways to compute $V^\pi(s)$. One approach uses a discount rate $\gamma$ to discount rewards over time. The sum is then computed over an infinite horizon:

$$V^\pi(s_t) = \sum_{i=0}^{\infty} \gamma^i r_{t+i} \tag{2}$$

where $r_t$ is the reward received at time step t. Alternatively, $V^\pi(s)$ could be computed by summing the rewards over a finite horizon $h$:

$$V^\pi(s_t) = \sum_{i=0}^{h} r_{t+i} \tag{3}$$

The agent's state descriptions are usually identified with the values returned by its *sensors*, which provide a description of both the agent's current state and the state of the world. Often the sensors do not give the agent complete state information and thus the state is only *partially observable*.

Besides reinforcement learning, intelligent agents can be designed by other paradigms, notably *planning* and *supervised learning*. We briefly note some of the major differences among these approaches. In general, planning methods require an explicit model of the state transition function $\delta(s, a)$. Given such a model, a planning algorithm can search through possible action choices to find an action sequence that will guide the agent from an initial state to a goal state. Since planning algorithms operate using a model of the environment, they can backtrack or "undo" state transitions that enter undesirable states. In contrast, RL is intended to apply to situations in which a sufficiently tractable action model does not exist. Consequently, an agent in the RL paradigm must actively explore its environment in order to observe the effects of its actions. Unlike planning, RL agents cannot normally undo state transitions. Of course, in some cases it may be possible to build up an action model through experience (Sutton, 1990), enabling more planning as experience accumulates. However, RL research focuses on the behavior of an agent when it has insufficient knowledge to perform planning.

Agents can also be trained through supervised learning. In supervised learning, the agent is presented with examples of state-action pairs, along with an indication that the action was either correct or incorrect. The goal in supervised learning is to induce a general policy from the training examples. Thus, supervised learning requires an *oracle* that can supply correctly labeled examples. In contrast, RL does not require prior knowledge of correct and incorrect decisions. RL can be applied to situations in which rewards are sparse; for example, rewards may be associated only with certain states. In such cases, it may be impossible to associate a label of "correct" or "incorrect" on particular decisions without reference to the agent's subsequent decisions, making supervised learning infeasible.

In summary, RL provides a flexible approach to the design of intelligent agents in situations for which both planning and supervised learning are impractical. RL can be applied to problems for which significant domain knowledge is either unavailable or costly to obtain. For example, a common RL task is robot control. Designers of autonomous robots often lack sufficient knowledge of the intended operational environment to use either the planning or the supervised learning regime to design a control policy for the robot. In this case, the goal of RL would be to enable the robot to generate effective decision policies as it explores its environment.

Figure 1 shows a simple sequential decision task that will be used as an example later in this paper. The task of the agent in this grid world is to move from state to state by selecting among two actions: right ($R$) or down ($D$). The sensor of the agent returns the identity of the current state. The agent always starts in state $a1$ and receives the reward indicated upon visiting each state. The task continues until the agent moves off the grid world (e.g., by taking action $D$ from state $a5$). The goal is to learn a policy that returns the highest cumulative rewards. For example, a policy which results in the sequences of actions $R, D, R, D, D, R, R, D$ starting from from state $a1$ gives the optimal score of 17.

|   | a | b | c | d | e |
|---|---|---|---|---|---|
| 1 | 0 | 2 | 1 | -1 | 1 |
| 2 | 1 | 1 | 2 | 0 | 2 |
| 3 | 3 | -5 | 4 | 3 | 1 |
| 4 | 1 | -2 | 4 | 1 | 2 |
| 5 | 1 | 1 | 2 | 1 | 1 |

Figure 1: A simple grid-world sequential decision task. The agent starts in state $a1$ and receives the row and column of the current box as sensory input. The agent moves from one box to another by selecting between two moves (right or down), and the agent's score is increased by the payoff indicated in each box. The goal is to find a policy that maximizes the cumulative score.

## 2.1 Policy Space vs. Value-Function Space

Given the reinforcement learning problem as described in the previous section, we now address the main topic: how to find an optimal policy, $\pi^*$. We consider two main approaches, one involves search in *policy space* and the other involves search in *value function space*.

Policy-space search methods maintain explicit representations of policies and modify them through a variety of search operators. Many search methods have been considered, including dynamic programming, value iteration, simulated annealing, and evolutionary algorithms. This paper focuses on evolutionary algorithms that have been specialized for the reinforcement learning task.

In contrast, value function methods do not maintain an explicit representation of a policy. Instead, they attempt learn the value function $V^{\pi^*}$, which returns the expected cumulative reward for the optimal policy from any state. The focus of research on value function approaches to RL is to design algorithms that learn these value functions through experience. The most common approach to learning value functions is the temporal difference (TD) method, which is described in the next section.

## 3. Temporal Difference Algorithms for Reinforcement Learning

As stated in the Introduction, a comprehensive comparison of value function search and direct policy-space search is beyond the scope of this paper. Nevertheless, it will be useful to point out key conceptual differences between typical value function methods and typical evolutionary algorithms for searching policy space. The most common approach for learning a value function $V$ for RL problems is the temporal difference (TD) method (Sutton, 1988).

The TD learning algorithm uses observations of *prediction differences* from consecutive states to update value predictions. For example, if two consecutive states $i$ and $j$ return payoff prediction values of 5 and 2, respectively, then the difference suggests that the payoff from state $i$ may be overestimated and should be reduced to agree with predictions from state $j$. Updates to the value function $V$ are achieved using the following update rule:

$$V(s_t) = V(s_t) + \alpha(V(s_{t+1}) - V(s_t) + r_t) \qquad (4)$$

where $\alpha$ represents the learning rate and $r_t$ any immediate reward. Thus, the difference in predictions $(V(s_{t+1}) - V(s_t))$ from consecutive states is used as a measure of prediction error. Consider a chain of value predictions $V(s_0)..V(s_n)$ from consecutive state transitions with the last prediction $V(s_n)$ containing the only non-zero reward from the environment. Over many iterations of this sequence, the update rule will adjust the values of each state so that they agree with their successors and eventually with the reward received in $V(s_n)$. In other words, the single reward is propagated backwards through the chain of value predictions. The net result is an accurate value function that can be used to predict the expected reward from any state of the system.

As mentioned earlier, the goal of TD methods is to learn the value function for the optimal policy, $V^{\pi^*}$. Given $V^{\pi^*}$, the optimal action, $\pi(s)$, can be computed using the following equation:

$$\pi(s) = \operatorname*{argmax}_a V^{\pi^*}(\delta(s,a)) \qquad (5)$$

Of course, we have already stated that in RL the state transition function $\delta(s,a)$ is unknown to the agent. Without this knowledge, we have no way of evaluating (5). An alternative value function that can be used to compute $\pi^*(s)$ is called a $Q$-function, $Q(s,a)$ (Watkins, 1989; Watkins & Dayan, 1992). The $Q$-function is a value function that represents the expected value of taking action $a$ in state $s$ and acting optimally thereafter:

$$Q(s,a) = r(s) + V^{\pi}(\delta(s,a)) \qquad (6)$$

where $r(s)$ represents any immediate reward received in state $s$. Given the $Q$-function, actions from the optimal policy can be directly computed using the following equation:

$$\pi^*(s) = \operatorname*{argmax}_a Q(s,a) \qquad (7)$$

Table 1 shows the $Q$-function for the grid world problem of Figure 1. This table-based representation of the $Q$-function associates cumulative future payoffs for each state-action pair in the system. (The letter-number pairs at the top represent the state given by the row and column in Figure 1, and $R$ and $D$ represent the actions *right* and *down*, respectively.) The TD method adjusts the $Q$-values after each decision. When selecting the next action, the agent considers the effect of that action by examining the expected value of the state transition caused by the action.

The $Q$-function is learned through the following TD update equation:

$$Q(s_t, a_t) = Q(s_t, a_t) + \alpha(\max_{a_{t+1}} Q(s_{t+1}, a_{t+1}) - Q(s_t, a_t) + r(s_t)) \qquad (8)$$

|   | a1 | a2 | a3 | a4 | a5 | b1 | b2 | b3 | b4 | b5 | c1 | c2 | c3 | c4 | c5 | d1 | d2 | d3 | d4 | d5 | e1 | e2 | e3 | e4 | e5 |
|---|----|----|----|----|----|----|----|----|----|----|----|----|----|----|----|----|----|----|----|----|----|----|----|----|----|
| R | 17 | 16 | 10 | 7  | 6  | 17 | 15 | 7  | 6  | 5  | 7  | 9  | 11 | 8  | 4  | 6  | 6  | 7  | 4  | 2  | 1  | 2  | 1  | 2  | 1  |
| D | 16 | 11 | 10 | 7  | 1  | 17 | 8  | 1  | 3  | 1  | 15 | 14 | 12 | 8  | 2  | 6  | 7  | 7  | 3  | 1  | 7  | 6  | 4  | 3  | 1  |

Table 1: A $Q$-function for the simple grid world. A value is associated with each state-action pair.

Essentially, this equation updates $Q(s_t, a_t)$ based on the current reward and the predicted reward if all future actions are selected optimally. Watkins and Dayan (1992) proved that if updates are performed in this fashion and if every $Q$-value is explicitly represented, the estimates will asymptotically converge to the correct values. A reinforcement learning system can thus use the $Q$ values to select the optimal action in any state. Because $Q$-learning is the most widely known implementation of temporal difference learning, we will use it in our qualitative comparisons with evolutionary approaches in later sections.

## 4. Evolutionary Algorithms for Reinforcement Learning (EARL)

The policy-space approach to RL searches for policies that optimize an appropriate objective function. While many search algorithms might be used, this survey focuses on evolutionary algorithms. We begin with a brief overview of a simple EA for RL, followed by a detailed discussion of features that characterize the general class of EAs for RL.

### 4.1 Design Considerations for Evolutionary Algorithms

Evolutionary algorithms (EAs) are global search techniques derived from Darwin's theory of evolution by natural selection. An EA iteratively updates a population of potential solutions, which are often encoded in structures called *chromosomes*. During each iteration, called a *generation*, the EA evaluates solutions and generates offspring based on the fitness of each solution in the task environment. Substructures, or *genes*, of the solutions are then modified through genetic operators such as mutation and recombination. The idea is that structures that are associated with good solutions can be mutated or combined to form even better solutions in subsequent generations. The canonical evolutionary algorithm is shown in Figure 2. There have been a wide variety of EAs developed, including genetic algorithms (Holland, 1975; Goldberg, 1989), evolutionary programming (Fogel, Owens, & Walsh, 1966), genetic programming (Koza, 1992), and evolutionary strategies (Rechenberg, 1964).

EAs are general purpose search methods and have been applied in a variety of domains including numerical function optimization, combinatorial optimization, adaptive control, adaptive testing, and machine learning. One reason for the widespread success of EAs is that there are relatively few requirements for their application, namely,

1. An appropriate mapping between the search space and the space of chromosomes, and

2. An appropriate fitness function.

```
procedure EA
begin
    t = 0;
    initialize P(t);
    evaluate structures in P(t);
    while termination condition not satisfied do
        begin
            t = t + 1;
            select P(t) from P(t-1);
            alter structures in P(t);
            evaluate structures in P(t);
        end
end.
```

Figure 2: Pseudo-code Evolutionary Algorithm.

For example, in the case of parameter optimization, it is common to represent the list of parameters as either a vector of real numbers or a bit string that encodes the parameters. With either of these representations, the "standard" genetic operators of mutation and cut-and-splice crossover can be applied in a straightforward manner to produce the genetic variations required (see Figure 3). The user must still decide on a (rather large) number of control parameters for the EA, including population size, mutation rates, recombination rates, parent selection rules, but there is an extensive literature of studies which suggest that EAs are relatively robust over a wide range of control parameter settings (Grefenstette, 1986; Schaffer, Caruana, Eshelman, & Das, 1989). Thus, for many problems, EAs can be applied in a relatively straightforward manner.

However, for many other applications, EAs need to be specialized for the problem domain (Grefenstette, 1987). The most critical design choice facing the user is the representation, that is, the mapping between the search space of knowledge structures (or, the *phenotype* space) and the space of chromosomes (the *genotype* space). Many studies have shown that the effectiveness of EAs is sensitive to the choice of representations. It is not sufficient, for example, to choose an arbitrary mapping from the search space into the space of chromosomes, apply the standard genetic operators and hope for the best. What makes a good mapping is a subject for continuing research, but the general consensus is that candidate solutions that share important phenotypic similarities must also exhibit similar forms of "building blocks" when represented as chromosomes (Holland, 1975). It follows that the user of an EA must carefully consider the most natural way to represent the elements of the search space as chromosomes. Moreover, it is often necessary to design appropriate mutation and recombination operators that are specific to the chosen representation. The end result of this design process is that the representation and genetic operators selected for the EA comprise a form of search *bias* similar to biases in other machine learning meth-ods. Given the proper bias, the EA can quickly identify useful "building blocks" within the population, and converge on the most promising areas of the search space.[1]

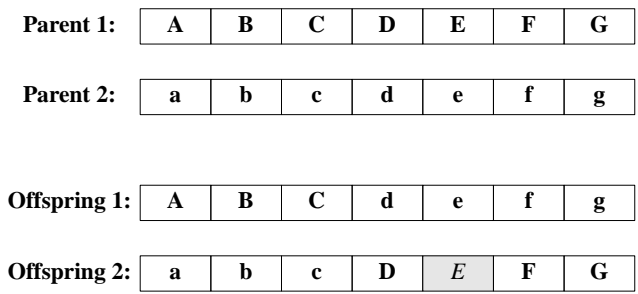

Figure 3: Genetic operators on fixed-position representation. The two offspring are generated by crossing over the selected parents. The operation shown is called *one-point crossover*. The first offspring inherits the initial segment of one parent and the final segment of the other parent. The second offspring inherits the same pattern of genes from the opposite parents. The crossover point is position 3, chosen at random. The second offspring has also incurred a mutation in the shaded gene.

In the case of RL, the user needs to make two major design decisions. First, how will the space of policies be represented by chromosomes in the EA? Second, how will the fitness of population elements be assessed? The answers to these questions depend on how the user chooses to bias the EA. The next section presents a simple EARL that adopts the most straightforward set of design decisions. This example is meant only to provide a baseline for comparison with more elaborate designs.

## 4.2 A Simple EARL

As the remainder of this paper shows, there are many ways to use EAs to search the space of RL policies. This section provides a concrete example of a simple EARL, which we call EARL$_1$. The pseudo-code is shown in Figure 4. This system provides the EA counterpart to the simple table-based TD system described in Section 3.

The most straightforward way to represent a policy in an EA is to use a single chromosome per policy with a single gene associated with each observed state. In EARL$_1$, each gene's value (or *allele* in biological terminology) represents the action value associated with the corresponding state, as shown in Figure 5. Table 2 shows part of an EARL$_1$ population of policies for the sample grid world problem. The number of policies in a population is usually on the order of 100 to 1000.

The fitness of each policy in the population must reflect the expected accumulated fitness for an agent that uses the given policy. There are no fixed constraints on how the fitness of an individual policy is evaluated. If the world is deterministic, like the sample grid-world,

---

1. Other ways to exploit problem specific knowledge in EAs include the use of heuristics to initialize the population and the hybridization with problem specific search algorithms. See (Grefenstette, 1987) for further discussions of these methods.

```
procedure EARL-1
begin
    t = 0;
    initialize a population of policies, P(t);
    evaluate policies in P(t);
    while termination condition not satisfied do
        begin
            t = t + 1;
            select high-payoff policies, P(t), from policies in P(t-1);
            update policies in P(t);
            evaluate policies in P(t);
        end
end.
```

Figure 4: Pseudo-code for Evolutionary Algorithm Reinforcement Learning system.

| | $s_1$ | $s_1$ | $s_3$ | | $s_N$ |
|---|---|---|---|---|---|
| **Policy $i$:** | $a_1$ | $a_1$ | $a_3$ | ... | $a_N$ |

Figure 5: Table-based policy representation. Each observed state has a gene which indicates the preferred action for that state. With this representation, standard genetic operators such as mutation and crossover can be applied.

the fitness of a policy can be evaluated during a single trial that starts with the agent in the initial state and terminates when the agent reaches a terminal state (e.g., falls off the grid in the grid-world). In non-deterministic worlds, the fitness of a policy is usually averaged over a sample of trials. Other options include measuring the total payoff achieved by the agent after a fixed number of steps, or measuring the number of steps required to achieve a fixed level of payoff.

Once the fitness of all policies in the population has been determined, a new population is generated according to the steps in the usual EA (Figure 2). First, parents are selected for reproduction. A typical selection method is to probabilistically select individuals based on relative fitness:

$$\Pr(p_i) = \frac{Fitness(p_i)}{\sum_{j=1}^{n} Fitness(p_j)} \qquad (9)$$

where $p_i$ represents individual $i$ and $n$ is the total number of individuals. Using this selection rule, the expected number of offspring for a given policy is proportional to that policy's fitness. For example, a policy with average fitness might have a single offspring, whereas a policy with twice the average fitness would have two offspring.[2] Offspring are formed by cloning the selected parents. Then new policies are generated by applying the standard genetic operators of crossover and mutation to the clones, as shown in Figure 3. The process of generating new populations of strategies can continue indefinitely or can be terminated after a fixed number of generations or once an acceptable level of performance is achieved.

| Policy | a1 | a2 | a3 | a4 | a5 | b1 | b2 | b3 | b4 | b5 | c1 | c2 | c3 | c4 | c5 | d1 | d2 | d3 | d4 | d5 | e1 | e2 | e3 | e4 | e5 | Fitness |
|---|---|---|---|---|---|---|---|---|---|---|---|---|---|---|---|---|---|---|---|---|---|---|---|---|---|---|
| 1 | D | R | D | D | D | R | R | R | R | R | D | R | D | D | D | R | R | D | R | R | R | D | R | R | D | 8 |
| 2 | D | D | D | D | R | R | R | R | R | R | D | D | R | R | D | R | D | R | R | R | D | R | D | D | R | 9 |
| 3 | R | D | D | R | R | D | R | D | R | R | D | D | D | D | R | R | R | R | D | R | R | D | D | D | D | 17 |
| 4 | D | D | D | D | R | D | R | R | R | R | R | D | R | R | R | D | R | R | D | R | D | R | D | D | R | 11 |
| 5 | R | D | D | D | R | D | R | R | D | R | R | D | R | R | D | R | D | R | R | D | D | R | D | D | D | 16 |

Table 2: An EA population of five decision policies for the sample grid world. This simple policy representation specifies an action for each state of the world. The fitness corresponds to the payoffs that are accumulated using each policy in the grid world.

For simple RL problems such as the grid-world, EARL₁ may provide an adequate approach. In later sections, we will point out some ways in which even EARL₁ exhibits strengths that are complementary to TD methods for RL. However, as in the case of TD methods, EARL methods have been extended to handle the many challenges inherent in more realistic RL problems. The following sections survey some of these extensions, organized around three specific biases that distinguish EAs for Reinforcement Learning (EARL) from more generic EAs: policy representations, fitness/credit-assignment models, and RL-specific genetic operators.

## 5. Policy Representations in EARL

Perhaps the most critical feature that distinguishes classes of EAs from one another is the representation used. For example, EAs for function optimization use a simple string or vector representation, whereas EAs for combinatorial optimization use distinctive representations for permutations, trees or other graph structures. Likewise, EAs for RL use a distinctive set of representations for policies. While the range of potential policy representations is unlimited, the representations used in most EARL systems to date can be largely categorized along two discrete dimensions. First, policies may be represented either by condition-action rules or by neural networks. Second, policies may be represented by a single chromosome or the representation may be distributed through one or more populations.

### 5.1 Single-Chromosome Representation of Policies

#### 5.1.1 RULE-BASED POLICIES

For most RL problems of practical interest, the number of observable states is very large, and the simple table-based representation in EARL₁ is impractical. For large scale state spaces, it is more reasonable to represent a policy as a set of condition-action rules in which the condition expresses a predicate that matches a set of states, as shown in Figure 6. Early examples of this representation include the systems LS-1 (Smith, 1983) and LS-2 (Schaffer & Grefenstette, 1985), followed later by SAMUEL (Grefenstette et al., 1990).

---

2. Many other parent selection rules have been explored (Grefenstette, 1997a, 1997b).

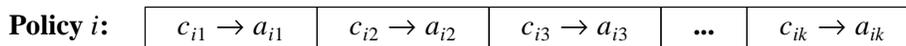

Figure 6: Rule-based policy representation. Each gene represents a condition-action rule that maps a set of states to an action. In general, such rules are independent of the position along the chromosome. Conflict resolution mechanisms may be needed if the conditions of rules are allowed to intersect.

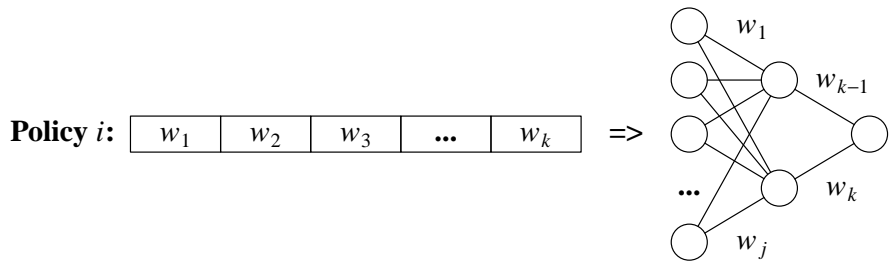

Figure 7: A simple parameter representation of weights for a neural network. The fitness of the policy is the payoff when the agent uses the corresponding neural net as its decision policy.

### 5.1.2 NEURAL NET REPRESENTATION OF POLICIES

As in TD-based RL systems, EARL systems often employ neural net representations as function approximators. In the simplest case (see Figure 7), a neural network for the agent's decision policy is represented as a sequence of real-valued connection weights. A straightforward EA for parameter optimization can be used to optimize the weights of the neural network (Belew, McInerney, & Schraudolph, 1991; Whitley, Dominic, Das, & Anderson, 1993; Yamauchi & Beer, 1993). This representation thus requires the least modification of the standard EA. We now turn to distributed representations of policies in EARL systems.

## 5.2 Distributed Representation of Policies

In the previous section we outlined EARL approaches that treat the agent's decision policy as a single genetic structure that evolves over time. This section addresses EARL approaches that decompose a decision policy into smaller components. Such approaches have two potential advantages. First, they allow evolution to work at a more detailed level of the task, e.g., on specific subtasks. Presumably, evolving a solution to a restricted subtask should be easier than evolving a monolithic policy for a complex task. Second, decomposition permits the user to exploit background knowledge. The user might base the decomposition into subtasks on a prior analysis of the overall performance task; for example, it might be known that certain subtasks are mutually exclusive and can therefore be learned independently. The user might also decompose a complex task into subtasks such that certain components can be explicitly programmed while other components are learned.

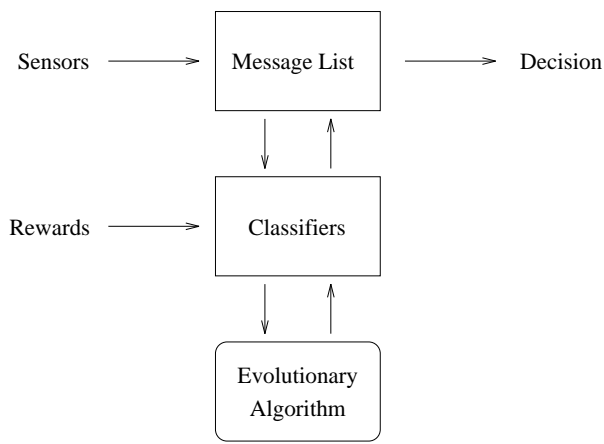

Figure 8: Holland's Learning Classifier System.

In terms of knowledge representation in EARL, the alternative to the single chromosome representation is to distribute the policy over several population elements. By assigning a fitness to these individual elements of the policy, evolutionary selection pressure can be brought to bear on more detailed aspects of the learning task. That is, fitness is now a function of individual subpolicies or individual rules or even individual neurons. This general approach is analogous to the classic TD methods that take this approach to the extreme of learning statistics concerning each state-action pair. As in the case of single-chromosome representations, we can partition distributed EARL representations into rule-based and neural-net-based classes.

### 5.2.1 DISTRIBUTED RULE-BASED POLICIES

The most well-known example of a distributed rule-based approach to EARL is the Learning Classifier Systems (LCS) model (Holland & Reitman, 1978; Holland, 1987; Wilson, 1994). An LCS uses an evolutionary algorithm to evolve if-then rules called *classifiers* that map sensory input to an appropriate action. Figure 8 outlines Holland's LCS framework (Holland, 1986). When sensory input is received, it is posted on the *message list*. If the left hand side of a classifier matches a message on the message list, its right hand side is posted on the message list. These new messages may subsequently trigger other classifiers to post messages or invoke a decision from the LCS, as in the traditional forward-chaining model of rule-based systems.

In an LCS, each chromosome represents a single decision rule and the entire population represents the agent's policy. In general, classifiers map a set of observed states to a set of messages, which may be interpreted as either internal state changes or actions. For example, if the learning agent for the grid world in Figure 1 has two sensors, one for the column and one for the row, then the population in an LCS might appear as shown in Table 3. The first classifier matches any state in the column *a* and recommends action *R*. Each classifier has a statistic called *strength* that estimates the utility of the rule. The strength statistics are used in both conflict resolution (when more than one action is recommended) and as fitness for the genetic algorithm. Genetic operators are applied to highly fit classifiers to generate new rules. Generally, the population size (i.e., the number of rules in the policy) is kept constant. Thus classifiers compete for space in the policy.

| condition | | action | strength |
|-----------|---|--------|----------|
| a#        | → | R      | 0.75     |
| #2        | → | D      | 0.25     |
|           | ... |      |          |
| d3        | → | D      | 0.50     |

Table 3: LCS population for grid world. The # is a *don't care* symbol which allows for generality in conditions. For example, the first rule says "Turn right in column *a*." The *strength* of a rule is used for conflict resolution and for parent selection in the genetic algorithm.

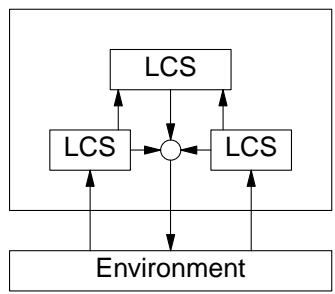

Figure 9: A two-level hierarchical ALECSYS system. Each LCS learns a specific behavior. The interactions among the rule sets are pre-programmed.

Another way that EARL systems distribute the representation of policies is to partition the policy into separate modules, with each module updated by its own EA. Dorigo and Colombetti (1998) describe an architecture called ALECSYS in which a complex reinforcement learning task is decomposed into subtasks, each of which is learned via a separate LCS, as shown in Figure 9. They provide a method called *behavior analysis and training* (BAT) to manage the incremental training of agents using the distributed LCS architecture.

The single-chromosome representation can also be extended by partitioning the policy across multiple co-evolving populations. For example, in the cooperative co-evolution model (Potter, 1997), the agent's policy is formed by combining chromosomes from several independently evolving populations. Each chromosome represents a set of rules, as in Figure 6, but these rules address only a subset of the performance task. For example, separate populations might evolve policies for different components of a complex task, or might address mutually exclusive sets of observed states. The fitness of each chromosome is computed based on the overall fitness of the agents that employ that chromosome as part of its combined chromosomes. The combined chromosomes represent the decision policy and are called a *collaboration* (Figure 10).

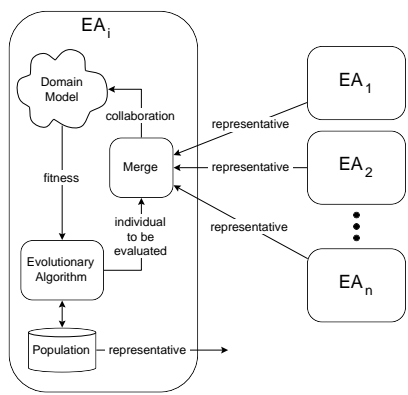

Figure 10: Cooperative coevolutionary architecture from the perspective of the $i^{th}$ EA instance. Each EA contributes a representative, which is merged with the others' representatives to form a *collaboration*, or policy for the agent. The fitness of each representative reflects the average fitness of its collaborations.

### 5.2.2 DISTRIBUTED NETWORK-BASED POLICIES

Distributed EARL systems using neural net representations have also been designed. In (Potter & De Jong, 1995), separate populations of neurons evolve, with the evaluation of each neuron based on the fitness of a collaboration of neurons selected from each population. In SANE (Moriarty & Miikkulainen, 1996a, 1998), two separate populations are maintained and evolved: a population of neurons and a population of network blueprints. The motivation for SANE comes from our *a priori* knowledge that individual neurons are fundamental building blocks in neural networks. SANE explicitly decomposes the neural network search problem into several parallel searches for effective single neurons. The neuron-level evolution provides evaluation and recombination of the neural network building blocks, while the population of blueprints search for effective combinations of these building blocks. Figure 11 gives an overview of the interaction of the two populations.

Each individual in the blueprint population consists of a set of pointers to individuals in the neuron population. During each generation, neural networks are constructed by combining the hidden neurons specified in each blueprint. Each blueprint receives a fitness according to how well the corresponding network performs in the task. Each neuron receives a fitness according to how well the top networks in which it participates perform in the task. An aggressive genetic selection and recombination strategy is used to quickly build and propagate highly fit structures in both the neuron and blueprint populations.

Network Blueprint Population          Neuron Population

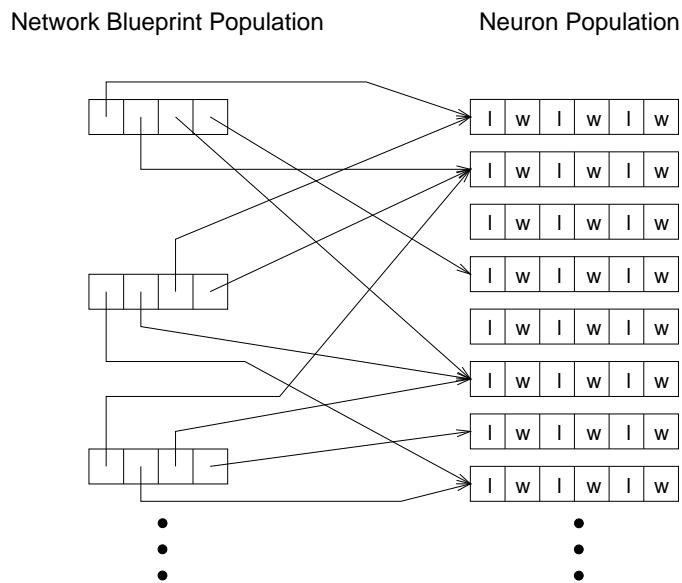

Figure 11: An overview of the two populations in SANE. Each member of the neuron population specifies a series of connections (connection labels and weights) to be made within a neural network. Each member of the network blueprint population specifies a series of pointers to specific neurons which are used to build a neural network.

## 6. Fitness and Credit Assignment in EARL

Evolutionary algorithms are all driven by the concept of natural selection: population elements that have higher fitness leave more offspring to later generations, thus influencing the direction of search in favor of high performance regions of the search space. The concept of fitness is central to any EA. In this section, we discuss features of the fitness model that are common across most EARL systems. We specifically focus on ways in which the fitness function reflects the distinctive structure of the RL problem.

### 6.1 The Agent Model

The first common features of all EARL fitness models is that fitness is computed with respect to an RL agent. That is, however the policy is represented in the EA, it must be converted to a decision policy for an agent operating in a RL environment. The agent is assumed to observe a description of the current state, select its next action by consulting its current policy, and collect whatever reward is provided by the environment. In EARL systems, as in TD systems, the agent is generally assumed to perform very little additional computation when selecting its next action. While neither approach limits the agent to strict stimulus-response behavior, it is usually assumed that the agent does not perform extensive planning or other reasoning before acting. This assumption reflects the fact that RL tasks involve some sort of control activity in which the agent must respond to a dynamic environment within a limited time frame.

## 6.2 Policy Level Credit Assignment

As shown in the previous section, the meaning of fitness in EARL systems may vary depending on what the population elements represent. In a single-chromosome representation, fitness is associated with entire policies; in a distributed representation, fitness may be associated with individual decision rules. In any case, fitness always reflects accumulated rewards received by the agent during the course of interaction with the environment, as specified in the RL model. Fitness may also reflect effort expended, or amount of delay.

It is worthwhile considering the different approaches to credit assignment in the TD and EA methods. In a reinforcement learning problem, payoffs may be sparse, that is, associated only with certain states. Consequently, a payoff may reflect the quality of an extended sequence of decisions, rather than any individual decision. For example, a robot may receive a reward after a movement that places it in a "goal" position within a room. The robot's reward, however, depends on many of its previous movements leading it to that point. A difficult *credit assignment* problem therefore exists in how to apportion the rewards of a sequence of decisions to individual decisions.

In general, EA and TD methods address the credit assignment problem in very different ways. In TD approaches, credit from the reward signal is explicitly propagated to each decision made by the agent. Over many iterations, payoffs are distributed across a sequence of decisions so that an appropriately discounted reward value is associated with each individual state and decision pair.

In simple EARL systems such as $\text{EARL}_1$, rewards are associated only with sequences of decisions and are not distributed to the individual decisions. Credit assignment for an individual decision is made implicitly, since policies that prescribe poor individual decisions will have fewer offspring in future generations. By selecting against poor policies, evolution automatically selects against poor individual decisions. That is, building blocks consisting of particular state-action pairs that are highly correlated with good policies are propagated through the population, replacing state-action pairs associated with poorer policies.

Figure 12 illustrates the differences in credit assignment between TD and $\text{EARL}_1$ in the grid world of Figure 1. The $Q$-learning TD method explicitly assigns credit or blame to each individual state-action pair by passing back the immediate reward and the estimated payoff from the new state. Thus, an error term becomes associated with each action performed by the agent. The EA approach does not explicitly propagate credit to each action but rather associates an overall fitness with the entire policy. Credit is assigned implicitly, based on the fitness evaluations of entire sequences of decisions. Consequently, the EA will tend to select against policies that generate the first and third sequences because they achieve lower fitness scores. The EA thus implicitly selects against action $D$ in state $b2$, for example, which is present in the bad sequences but not present in the good sequences.

## 6.3 Subpolicy Credit Assignment

Besides the implicit credit assignment performed on building blocks, EARL systems have also addressed the credit assignment problem more directly. As shown in Section 4, the individuals in an EARL system might represent either entire policies or components of a policy (e.g., component rule-sets, individual decision rules, or individual neurons). For distributed-representation EARLs, fitness is explicitly assigned to individual components.

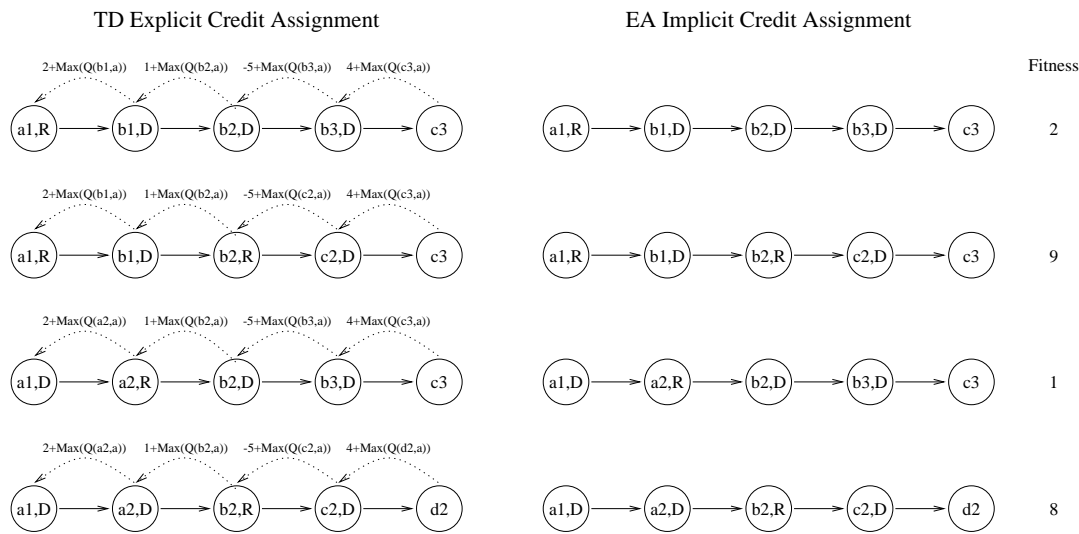

Figure 12: Explicit vs. implicit credit assignment. The $Q$-learning TD method assigns credit to each state-action pair based on the immediate reward and the predicted future rewards. The EA method assigns credit implicitly by associating fitness values with entire sequences of decisions.

In cases in which a policy is represented by explicit components, different fitness functions can be associated with different evolving populations, allowing the implementer to "shape" the overall policy by evolving subpolicies for specific subtasks (Dorigo & Colombetti, 1998; Potter, De Jong, & Grefenstette, 1995). The most ambitious goal is to allow the system to manage the number of co-evolving species as well as the form of interactions (Potter, 1997). This exciting research is still at an early stage.

For example, in the LCS model, each classifier (decision rule) has a *strength* which is updated using a TD-like method called the *bucket brigade algorithm* (Holland, 1986). In the bucket brigade algorithm, the strength of a classifier is used to bid against other classifiers for the right to post messages. Bids are subtracted from winning classifiers and passed back to the classifiers that posted the enabling message on the previous step. Classifier strengths are thus reinforced if the classifier posts a message that triggers another classifier. The classifier that invokes a decision from the LCS receives a strength reinforcement directly from the environment. The bucket brigade bid passing mechanism clearly bears a strong relation to the method of temporal differences (Sutton, 1988). The bucket brigade updates a given classifier's strength based on the strength of the classifiers that fire as a direct result of its activation. The TD methods differ slightly in this respect because they assign credit based strictly on temporal succession and do not take into account causal relations of steps. It remains unclear which is more appropriate for distributing credit.

Even for single chromosome representations, TD-like methods have been adopted in some EARL systems. In Samuel, each gene (decision rule) also maintains a quantity called *strength* that is used to resolve conflict when more than one rule matches the agent's current sensor readings. When payoff is obtained (thereby terminating the *trial*), the strengths of all rules that fired during the trial are updated (Grefenstette, 1988). In addition to resolving conflicts, a rule's strength also plays a role in triggering mutation operations, as described in the next section.

## 7. RL-Specific Genetic Operators

The creation of special genetic operators provides another avenue for imposing an RL-specific bias on EAs. Specialized operators in EARL systems first appeared in (Holland, 1986), in which so-called *triggered operators* were responsible for creating new classifiers when the learning agent found that no classifier in its existing population matched the agent's current sensor readings. In this case, a high-strength rule was explicitly generalized to cover the new set of sensor readings. A similar rule-creation operator was included in early versions of Samuel (Grefenstette et al., 1990). Later versions of Samuel included a number of mutation operators which created altered rules based on an agent's early experiences. For example, Samuel's *Specialization* mutation operator is triggered when a low-strength, general rule fires during an episode that results in high payoff. In such a case, the rule's conditions are reduced in generality to more closely match the agent's sensor readings. For example, if the agent has a sensor readings ($range = 40$, $bearing = 100$) and the original rule is:

IF $range = [25, 55]$ AND $bearing = [0, 180]$ THEN SET $turn = 24$ ($strength$ 0.1)

then the new rule would be:

IF $range = [35, 45]$ AND $bearing = [50, 140]$ THEN SET $turn = 24$ ($strength$ 0.8)

Since the episode triggering the operator resulted in high payoff, one might suspect that the original rule was over-generalized, and that the new, more specific version might lead to better results. (The strength of the new rule is initialized to the payoff received during the triggering episode.) This is considered a Lamarckian operator because the agent's experience is causing a genetic change which is passed on to later offspring.[3]

Samuel also uses an RL-specific crossover operator to recombine policies. In particular, crossover in Samuel attempts to cluster decision rules before assigning them to offspring. For example, suppose that the traces of the most previous evaluations of the parent strategies are as follows ($R_{i,j}$ denotes the $j^{\text{th}}$ decision rule in policy $i$):

Trace for parent #1:
Episode:

$\vdots$

8. $R_{1,3} \rightarrow R_{1,1} \rightarrow R_{1,7} \rightarrow R_{1,5}$     High Payoff
9. $R_{1,2} \rightarrow R_{1,8} \rightarrow R_{1,4}$     Low Payoff

---

3. Jean Baptiste Lamarck developed an evolutionary theory that stressed the inheritance of acquired characteristics, in particular acquired characteristics that are well adapted to the surrounding environment. Of course, Lamarck's theory was superseded by Darwin's emphasis on two-stage adaptation: undirected variation followed by selection. Research has generally failed to substantiate any Lamarckian mechanisms in biological systems (Gould, 1980).

$\vdots$

Trace for parent #2:

$\vdots$

4. $R_{2,7} \rightarrow R_{2,5}$                 Low Payoff

5. $R_{2,6} \rightarrow R_{2,2} \rightarrow R_{2,4}$       High Payoff

$\vdots$

Then one possible offspring would be:

$$\{R_{1,8}, \ldots, R_{1,3}, R_{1,1}, R_{1,7}, R_{1,5}, \ldots, R_{2,6}, R_{2,2}, R_{2,4}, \ldots, R_{2,7}\}$$

The motivation here is that rules that fire in sequence to achieve a high payoff should be treated as a group during recombination, in order to increase the likelihood that the offspring policy will inherit some of the better behavior patterns of its parents. Rules that do not fire in successful episodes (e.g., $R_{1,8}$) are randomly assigned to one of the two offspring. This form of crossover is not only Lamarckian (since it is triggered by the experiences of the agent), but is directly related to the structure of the RL problem, since it groups components of policies according to the temporal association among the decision rules.

## 8. Strengths of EARL

The EA approach represents an interesting alternative for solving RL problems, offering several potential advantages for scaling up to realistic applications. In particular, EARL systems have been developed that address difficult challenges in RL problems, including:

- Large state spaces;

- Incomplete state information; and

- Non-stationary environments.

This section focuses on ways that EARL address these challenges.

### 8.1 Scaling Up to Large State Spaces

Many early papers in the RL literature analyze the efficiency of alternative learning methods on toy problems similar to the grid world shown in Figure 1. While such studies are useful as academic exercises, the number of observed states in realistic applications of RL is likely to preclude any approach that requires the explicit storage and manipulation of statistics associated with each observable state-action pair. There are two ways that EARL policy representations help address the problem of large state spaces: *generalization* and *selectivity*.

#### 8.1.1 POLICY GENERALIZATION

Most EARL policy representations specify the policy at a level of abstraction higher than an explicit mapping from observed states to actions. In the case of rule-based representations, the rule language allows conditions to match sets of states, thus greatly reducing the storage required to specify a policy. It should be noted, however, that the generality of the rules within a policy may vary considerably, from the level of rules that specify an action for a single observed state all the way to completely general rules that recommend an action regardless of the current state. Likewise, in neural net representations, the mapping function is stored implicitly in the weights on the connections of the neural net. In either case, a generalized policy representation facilitates the search for good policies by grouping together states for which the same action is required.

| | a1 | a2 | a3 | a4 | a5 | b1 | b2 | b3 | b4 | b5 | c1 | c2 | c3 | c4 | c5 | d1 | d2 | d3 | d4 | d5 | e1 | e2 | e3 | e4 | e5 |
|---|---|---|---|---|---|---|---|---|---|---|---|---|---|---|---|---|---|---|---|---|---|---|---|---|---|
| R | 16 | 7 | ? | 17 | 12 | 8 | 12 | 11 | 11 | 12 | 14 | 7 | 12 | 13 | 9 | 12 | 11 | 12 | 12 | 11 | ? | 12 | 7 | ? | 9 |
| L | 9 | 13 | 12 | 11 | ? | 15 | ? | 17 | 16 | ? | 11 | 13 | 12 | 7 | 14 | 11 | 12 | ? | 11 | 16 | 12 | ? | 13 | 12 | 16 |

Table 4: An approximated value function from the population in Table 2. The table displays the average fitness for policies that select each state-action pair and reflects the estimated impact each action has on overall fitness. Given the tiny population size in this example, the estimates are not particularly accurate. Note the question marks in states where actions have converged. Since no policies select the alternative action, the population has no statistics on the impact of these actions on fitness. This is different from simple TD methods, where statistics on all actions are maintained.

### 8.1.2 Policy Selectivity

Most EARL systems have *selective* representations of policies. That is, the EA learns mappings from observed states to recommended actions, usually eliminating explicit information concerning less desirable actions. Knowledge about bad decisions is not explicitly preserved, since policies that make such decisions are selected against by the evolutionary algorithm and are eventually eliminated from the population. The advantage of selective representations is that attention is focused on profitable actions only, reducing space requirements for policies.

Consider our example of the simple EARL operating on the grid world. As the population evolves, policies normally converge to the best actions from a specific state, because of the selective pressure to achieve high fitness levels. For example, the population shown in Table 2 has converged alleles (actions) in states $a3, a5, b2, b5, d3, e1$, and $e2$. Each of these converged state-action pairs is highly correlated with fitness. For example, all policies have converged to action $R$ in state $b2$. Taking action $R$ in state $b2$ achieves a much higher expected return than action $D$ (15 *vs.* 8 from Table 1). Policies that select action $D$ from state $b2$ achieve lower fitness scores and are selected against. For this simple EARL, a snapshot of the population (Table 2) provides an implicit estimate of a corresponding TD value function (Table 4), but the distribution is biased toward the more profitable state-actions pairs.

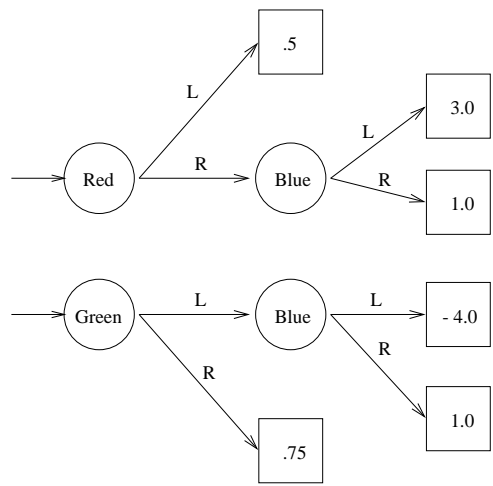

Figure 13: An environment with incomplete state information. The circles represent the states of the world and the colors represent the agent's sensory input. The agent is equally likely to start in the *red* state or the *green* state

## 8.2 Dealing with Incomplete State Information

Clearly, the most favorable condition for reinforcement learning occurs when the agent can observe the true state of the dynamic system with which it interacts. When complete state information is available, TD methods make efficient use of available feedback by associating reward directly with individual decisions. In real world situations, however, the agent's sensors are more likely to provide only a partial view that may fail to disambiguate many states. Consequently, the agent will often be unable to completely distinguish its current state. This problem has been termed *perceptual aliasing* or the *hidden state* problem. In the case of limited sensory information, it may be more useful to associate rewards with larger blocks of decisions. Consider the situation in Figure 13, in which the agent must act without complete state information. Circles represent the specific states of the world, and the colors represent the sensor information the agent receives within the state. Square nodes represent goal states with the corresponding reward shown inside. In each state, the agent has a choice of two actions ($L$ or $R$). We further assume that the state transitions are deterministic and that the agent is equally likely to start in either the state with the red or green sensor readings.

In this example, there are two different states that return a sensor reading of *blue*, and the agent is unable to distinguish between them. Moreover, the actions for each *blue* state return very different rewards. A $Q$ function applied to this problem treats the sensor reading of *blue* as one observable state, and the rewards for each action are averaged over both *blue* states. Thus, $Q(blue, L)$ and $Q(blue, R)$ will converge to -0.5 and 1, respectively. Since the reward from $Q(blue, R)$ is higher than the alternatives from observable states *red* and *green*, the agent's policy under Q-learning will choose to enter observable state *blue* each time. The final decision policy under Q-learning is shown in Table 5. This table also shows the optimal policy with respect to the agent's limited view of its world. In other

|         | Value Function Policy | Optimal Policy |
|---------|:---------------------:|:--------------:|
| *Red*   | R                     | R              |
| *Green* | L                     | R              |
| *Blue*  | R                     | L              |
| Expected Reward | 1.0           | 1.875          |

Table 5: The policy and expected reward returned by a converged $Q$ function compared to the optimal policy given the same sensory information.

words, the policy reflects the optimal choices if the agent cannot distinguish the two blue states.

By associating values with individual observable states, the simple TD methods are vulnerable to hidden state problems. In this example, the ambiguous state information misleads the TD method, and it mistakenly combines the rewards from two different states of the system. By confounding information from multiple states, TD cannot recognize that advantages might be associated with specific actions from specific states, for example, that action $L$ from the top *blue* state achieves a very high reward.

In contrast, since EA methods associate credit with entire policies, they rely more on the net results of decision sequences than on sensor information, that may, after all, be ambiguous. In this example, the evolutionary algorithm exploits the disparity in rewards from the different *blue* states and evolves policies that enter the good *blue* state and avoid the bad one. The agent itself remains unable to distinguish the two *blue* states, but the evolutionary algorithm implicitly distinguishes among ambiguous states by rewarding policies that avoid the bad states.

For example, an EA method can be expected to evolve an optimal policy in the current example given the existing, ambiguous state information. Policies that choose the action sequence $R,L$ when starting in the *red* state will achieve the highest levels of fitness, and will therefore be selected for reproduction by the EA. If agents using these policies are placed in the *green* state and select action $L$, they receive the lowest fitness score, since their subsequent action, $L$ from the *blue* sensors, returns a negative reward. Thus, many of the policies that achieve high fitness when started in the *red* state will be selected against if they choose $L$ from the *green* state. Over the course of many generations, the policies must choose action $R$ from the *green* state to maximize their fitness and ensure their survival.

We confirmed these hypotheses in empirical tests. A $Q$-learner using single-step updates and a table-based representation converged to the values in Table 5 in every run. An evolutionary algorithm[4] consistently converged 80% of its population on the optimal policy. Figure 14 shows the average percentage of the optimal policy in the population as a function of time, averaged over 100 independent runs.

Thus even simple EA methods such as EARL$_1$ appear to be more robust in the presence of hidden states than simple TD methods. However, more refined sensor information could still be helpful. In the previous example, although the EA policies achieve a better average reward than the TD policy, the evolved policy remains unable to procure both the 3.0 and 1.0 rewards from the two *blue* states. These rewards could be realized, however, if the agent could separate the two *blue* states. Thus, any method that generates additional features to disambiguate states presents an important asset to EA methods. Kaelbling et al. (1996) describe several promising solutions to the hidden state problem, in which additional features such as the agent's previous decisions and observations are automatically generated and included in the agent's sensory information (Chrisman, 1992; Lin & Mitchell, 1992; McCallum, 1995; Ring, 1994). These methods have been effective at disambiguating states for TD methods in initial studies, but further research is required to determine the extent to which similar methods can resolve significant hidden state information in realistic applications. It would be useful to develop ways to use such methods to augment the sensory data available in EA methods as well.

---

4. We used a binary tournament selection, a 50 policy population, 0.8 crossover probability, and 0.01 mutation rate.

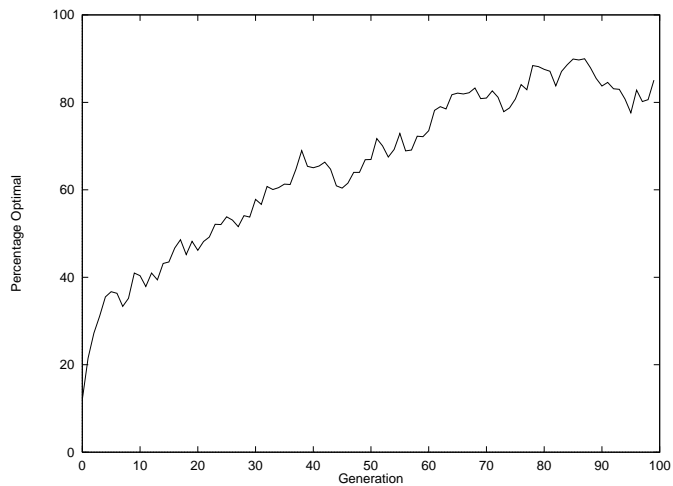

Figure 14: The optimal policy distribution in the hidden state problem for an evolutionary algorithm. The graph plots the percentage of optimal policies in the population, averaged over 100 runs.

## 8.3 Non-Stationary Environments

If the agent's environment changes over time, the RL problem becomes even more difficult, since the optimal policy becomes a moving target. The classic trade-off between exploration and exploitation becomes even more pronounced. Techniques for encouraging exploration in TD-based RL include adding an *exploration bonus* to the estimated value of state-action pairs that reflects how long it has been since the agent has tried that action (Sutton, 1990), and building a statistical model of the agent's uncertainty (Dayan & Sejnowski, 1996). Simple modifications of standard evolutionary algorithms offer an ability to track non-stationary environments, and thus provide a promising approach to RL for these difficult cases.

The fact that evolutionary search is based on competition within a population of policies suggest some immediate benefits for tracking non-stationary environments. To the extent that the population maintains a diverse set of policies, changes in the environment will bias selective pressure in favor of the policies that are most fit for the current environment. As long as the environment changes slowly with respect to the time required to evaluate a population of policies, the population should be able to track a changing fitness landscape without any alteration of the algorithm. Empirical studies show that maintaining the diversity within the population may require a higher mutation rate than those usually adopted for stationary environments (Cobb & Grefenstette, 1993).

In addition, special mechanisms have been explored in order to make EAs more responsive to rapidly changing environments. For example, (Grefenstette, 1992) suggests maintaining a random search within a restricted portion of the population. The random population elements are analogous to immigrants from other populations with uncorrelated fitness landscapes. Maintaining this source of diversity permits the EA to respond rapidly to large, sudden changes in the fitness landscape. By keeping the randomized portion of the population to less than about 30% of the population, the impact on search efficiency in stationary environments is minimized. This is a general approach that can easily be applied in EARL systems.

Other useful algorithms that have been developed to ensure diversity in evolving populations include fitness sharing (Goldberg & Richardson, 1987), crowding (De Jong, 1975), and local mating (Collins & Jefferson, 1991). In Goldberg's fitness sharing model, for example, similar individuals are forced to share a large portion of a single fitness value from the shared solution point. Sharing decreases the fitness of similar individuals and causes evolution to select against individuals in overpopulated niches.

EARL methods that employ distributed policy representations achieve diversity automatically and are well-suited for adaptation in dynamic environments. In a distributed representation, each individual represents only a partial solution. Complete solutions are built by combining individuals. Because no individual can solve the task on its own, the evolutionary algorithm will search for several complementary individuals that together can solve the task. Evolutionary pressures are therefore present to prevent convergence of the population. Moriarty and Miikkulainen (1998) showed how the inherent diversity and specialization in SANE allow it to adapt much more quickly to changes in the environment than standard, convergent evolutionary algorithms.

Finally, if the learning system can detect changes in the environment, even more direct response is possible. In the *anytime learning* model (Grefenstette & Ramsey, 1992), an EARL system maintains a case-base of policies, indexed by the values of the environmental detectors corresponding to the environment in which a given policy was evolved. When an environmental change is detected, the population of policies is partially reinitialized, using previously learned policies selected on the basis of similarity between the previously encountered environment and the current environment. As a result, if the environment changes are cyclic, then the population can be immediately seeded with those policies in effect during the last occurrence of the current environment. By having a population of policies, this approach is protected against some kinds of errors in detecting environmental changes. For example, even if a spurious environmental change is mistakenly detected, learning is not unduly affected, since only a part of the current population of policies is replaced by previously learned policies. Zhou (1990) explored a similar approach based on LCS.

In summary, EARL systems can respond to non-stationary environments, both by techniques that are generic to evolutionary algorithms and by techniques that have been specifically designed with RL in mind.

## 9. Limitations of EARL

Although the EA approach to RL is promising and has a growing list of successful applications (as outlined in the following section), a number of challenges remain.

### 9.1 Online Learning

We can distinguish two broad approaches to reinforcement learning —*online learning* and *offline learning*. In online learning, an agent learns directly from its experiences in its operational environment. For example, a robot might learn to navigate in a warehouse by actually moving about its physical environment. There are two problems with using EARL in this situation. First, it is likely to require a large number of experiences in order to evaluate a large population of policies. Depending on how quickly the agent performs tasks that result in some environmental feedback, it may take an unacceptable amount of time to run hundreds of generations of an EA that evaluates hundreds or thousands of policies. Second, it may be dangerous or expensive to permit an agent to perform some actions in its actual operational environment that might cause harm to itself or its environment. Yet it is very likely that at least some policies that the EA generates will be very bad policies. Both of these objections apply to TD methods as well. For example, the theoretical results that prove the optimality of Q-learning require that every state be visited infinitely often, which is obviously impossible in practice. Likewise, TD methods may explore some very undesirable states before an acceptable value-function is found.

For both TD and EARL, practical considerations point toward the use of offline learning, in which the RL system performs its exploration on simulation models of the environment. Simulation models provide a number of advantages for EARL, including the ability to perform parallel evaluations of all the policies in a population simultaneously (Grefenstette, 1995).

### 9.2 Rare States

The memory or record of observed states and rewards differs greatly between EA and TD methods. Temporal difference methods normally maintain statistics concerning every state-action pair. As states are revisited, the new reinforcement is combined with the previous value. New information thus supplements previous information, and the information content of the agent's reinforcement model increases during exploration. In this manner, TD methods sustain knowledge of both good and bad state-action pairs.

As pointed out previously, EA methods normally maintain information only about good policies or policy components. Knowledge of bad decisions is not explicitly preserved, since policies that make such decisions are selected against by the evolutionary algorithm and are eventually eliminated from the population. For example, refer once again to Table 4, which shows the implicit statistics of the population from Table 2. Note the question marks in states where actions have converged. Since no policies in the population select the alternative action, the EA has no statistics on the impact of these actions on fitness.

This reduction in information content within the evolving population can be a disadvantage with respect to states that are rarely visited. In any evolutionary algorithm, the value of genes that have no real impact on the fitness of the individual tends to drift to random values, since mutations tend to accumulate in these genes. If a state is rarely encountered, mutations may freely accumulate in the gene that describes the best action for that state. As a result, even if the evolutionary algorithm learns the correct action for a rare state, that information may eventually be lost due to mutations. In contrast, since table-based TD methods permanently record information about all state-action pairs, they may be more robust when the learning agent does encounter a rare state. Of course, if a TD method uses a function approximator such as a neural network as its value function, then it too can suffer from memory loss concerning rare states, since many updates from frequently occurring states can dominate the few updates from the rare states.

## 9.3 Proofs of Optimality

One of the attractive features of TD methods is that the Q-learning algorithm has a proof of optimality (Watkins & Dayan, 1992). However, the practical importance of this result is limited, since the assumptions underlying the proof (e.g., no hidden states, all state visited infinitely often) are not satisfied in realistic applications. The current theory of evolutionary algorithms provide a similar level of optimality proofs for restricted classes of search spaces (Vose & Wright, 1995). However, no general theoretical tools are available that can be applied to realistic RL problems. In any case, ultimate convergence to an optimal policy may be less important in practice than efficiently finding a reasonable approximation.

A more pragmatic approach may be to ask how efficient alternative RL algorithms are, in terms of the number of reinforcements received before developing a policy that is within some tolerance level of an optimal policy. In the model of *probably approximately correct* (PAC) learning (Valiant, 1984), the performance of a learner is measured by how many learning experiences (e.g., samples in supervised learning) are required before converging to a correct hypothesis within specified error bounds. Although developed initially for supervised learning, the PAC approach has been extended recently to both TD methods (Fiechter, 1994) and to general EA methods (Ros, 1997). These analytic methods are still in an early stage of development, but further research along these lines may one day provide useful tools for understanding the theoretical and practical advantages of alternative approaches to RL. Until that time, experimental studies will provide valuable evidence for the utility of an approach.

## 10. Examples of EARL Methods

Finally, we take a look at a few significant examples of the EARL approach and results on RL problems. Rather than attempt an exhaustive survey, we have selected four EARL systems that are representative of the diverse policies representations outlined in Section 5. SAMUEL represents the class of single-chromosome rule-based EARL systems. ALECSYS is an example of a distributed rule-based EARL method. GENITOR is a single chromosome neural-net system, and SANE is a distributed neural net system. This brief survey should provide a starting point for those interested in investigating the evolutionary approach to reinforcement learning.

## 10.1 SAMUEL

SAMUEL (Grefenstette et al., 1990) is an EARL system that combines Darwinian and Lamarckian evolution with aspects of temporal difference reinforcement learning. SAMUEL has been used to learn behaviors such as navigation and collision avoidance, tracking, and herding, for robots and other autonomous vehicles.

SAMUEL uses a single-chromosome, rule-based representation for policies, that is, each member of the population is a policy represented as a rule set and each gene is a rule that maps the state of the world to actions to be performed. An example rule might be:

IF $range = [35, 45]$ AND $bearing = [0, 45]$ THEN SET $turn = 16$ ($strength$ 0.8)

The use of a high-level language for rules offers several advantages over low-level binary pattern languages typically adopted in genetic learning systems. First, it makes it easier to incorporate existing knowledge, whether acquired from experts or by symbolic learning programs. Second, it is easier to transfer the knowledge learned to human operators. SAMUEL also includes mechanisms to allow coevolution of multiple behaviors simultaneously. In addition to the usual genetic operators of crossover and mutation, SAMUEL uses more traditional machine learning techniques in the form of Lamarckian operators. SAMUEL keeps a record of recent experiences and will allow operators such as generalization, specialization, covering, and deletion to make informed changes to the individual genes (rules) based on these experiences.

SAMUEL has been used successfully in many reinforcement learning applications. Here we will briefly describe three examples of learning complex behaviors for real robots. In these applications of SAMUEL, learning is performed under simulation, reflecting the fact that during the initial phases of learning, controlling a real system can be expensive or dangerous. Learned behaviors are then tested on the on-line system.

In (Schultz & Grefenstette, 1992; Schultz, 1994; Schultz & Grefenstette, 1996), SAMUEL is used to learn collision avoidance and local navigation behaviors for a Nomad 200 mobile robot. The sensors available to the learning task were five sonars, five infrared sensors, and the range and bearing to the goal, and the current speed of the vehicle. SAMUEL learned a mapping from those sensors to the controllable actions – a turning rate and a translation rate for the wheels. SAMUEL took a human-written rule set that could reach the goal within a limited time without hitting an obstacle only 70 percent of the time, and after 50 generations was able to obtain a 93.5 percent success rate.

In (Schultz & Grefenstette, 1996), the robot learned to herd a second robot to a "pasture". In this task, the learning system used the range and bearing to the second robot, the heading of the second robot, and the range and bearing to the goal, as its input sensors. The system learned a mapping from these sensors to a turning rate and steering rate. In these experiments, success was measured as the percentage of times that the robot could maneuver the second robot to the goal within a limited amount of time. The second robot implemented a random walk, plus a behavior that made it avoid any nearby obstacles. The first robot learned to exploit this to achieve its goal of moving the second robot to the goal.

SAMUEL was given an initial, human-designed rule set with a performance of 27 percent, and after 250 generations was able to move the second robot to the goal 86 percent of the time.

In (Grefenstette, 1996) the SAMUEL EA system is combined with case-based learning to address the adaptation problem. In this approach, called *anytime learning* (Grefenstette & Ramsey, 1992), the learning agent interacts both with the external environment and with an internal simulation. The anytime learning approach involves two continuously running and interacting modules: an execution module and a learning module. The execution module controls the agent's interaction with the environment and includes a monitor that dynamically modifies the internal simulation model based on observations of the actual agent and the environment. The learning module continuously tests new strategies for the agent against the simulation model, using a genetic algorithm to evolve improved strategies, and updates the knowledge base used by the execution module with the best available results. Whenever the simulation model is modified due to some observed change in the agent or the environment, the genetic algorithm is restarted on the modified model. The learning system operates indefinitely, and the execution system uses the results of learning as they become available. The work with SAMUEL shows that the EA method is particularly well-suited for anytime learning. Previously learned strategies can be treated as cases, indexed by the set of conditions under which they were learned. When a new situation is encountered, a nearest neighbor algorithm is used to find the most similar previously learned cases. These nearest neighbors are used to re-initialize the genetic population of policies for the new case. Grefenstette (1996) reports on experiments in which a mobile robot learns to track another robot, and dynamically adapts its policies using anytime learning as its encounters a series of partial system failures. This approach blurs the line between online and offline learning, since the online system is being updated whenever the offline learning system develops an improved policy. In fact, the offline learning system can even be executed on-board the operating mobile robot.

## 10.2 ALECSYS

As described previously, ALECSYS (Dorigo & Colombetti, 1998) is a distributed rule-based EA that supports an approach to the design of autonomous systems called *behavioral engineering*. In this approach, the tasks to be performed by a complex autonomous systems are decomposed into individual behaviors, each of which is learned via a learning classifier systems module, as shown in Figure 9. The decomposition is performed by the human designer, so the fitness function associated with each LCS can be carefully designed to reflect the role of the associated component behavior within the overall autonomous system. Furthermore, the interactions among the modules is also preprogrammed. For example, the designer may decide that the robot should learn to approach a goal except when a threatening predator is near, in which case the robot should evade the predator. The overall architecture of the set of behaviors can then be set such that the evasion behavior has higher priority than the goal-seeking behavior, but the individual LCS modules can evolve decision rules for optimally performing the subtasks.

ALECSYS has been used to develop behavioral rules for a number of behaviors for autonomous robots, including complex behavior groups such as CHASE/FEED/ESCAPE (Dorigo & Colombetti, 1998). The approach has been implemented and tested on both simulated robots and on real robots. Because it exploits both human design and EARL methods to optimize system performance, this method shows much promise for scaling up to realistic tasks.

## 10.3 Genitor

Genitor (Whitley & Kauth, 1988; Whitley, 1989) is an aggressive, general purpose genetic algorithm that has been shown effective when specialized for use on reinforcement-learning problems. Whitley et al. (1993) demonstrated how Genitor can efficiently evolve decision policies represented as neural networks using only limited reinforcement from the domain.

Genitor relies solely on its evolutionary algorithm to adjust the weights in neural networks. In solving RL problems, each member of the population in Genitor represents a neural network as a sequence of connection weights. The weights are concatenated in a real-valued chromosome along with a gene that represents a crossover probability. The crossover gene determines whether the network is to be mutated (randomly perturbed) or whether a crossover operation (recombination with another network) is to be performed. The crossover gene is modified and passed to the offspring based on the offspring's performance compared to the parent. If the offspring outperforms the parent, the crossover probability is decreased. Otherwise, it is increased. Whitley et al. refer to this technique as *adaptive mutation*, which tends to increase the mutation rate as populations converge. Essentially, this method promotes diversity within the population to encourage continual exploration of the solution space.

Genitor also uses a so-called "steady-state" genetic algorithm in which new parents are selected and genetic operators are applied after each individual is evaluated. This approach contrasts with "generational" GAs in which the entire population is evaluated and replaced during each generation. In a steady-state GA, each policy is evaluated just once and retains this same fitness value indefinitely. Since policies with lower fitness are more likely to be replaced, it is possible that a fitness based on a noisy evaluation function may have an undesirable influence on the direction of the search. In the case of the pole-balancing RL application, the fitness value depends on the length of time that the policy can maintain a good balance, given a randomly chosen initial state. The fitness is therefore a random variable that depends on the initial state. The authors believe that noise in the fitness function had little negative impact on learning good policies, perhaps because it was more difficult for poor networks to obtain a good fitness than for good networks (of which there were many copies in the population) to survive an occasional bad fitness evaluation. This is an interesting general issue in EARL that needs further analysis.

Genitor adopts some specific modification for its RL applications. First, the representation uses a real-valued chromosome rather than a bit-string representation for the weights. Consequently, Genitor always recombines policies between weight definitions, thus reducing potentially random disruption of neural network weights that might result if crossover operations occurred in the middle of a weight definition. The second modification is a very high mutation rate which helps to maintain diversity and promote rapid exploration of the policy space. Finally, Genitor uses unusually small populations in order to discourage different, competing neural network "species" from forming within the population. Whit-ley et al. (1993) argue that speciation leads to competing conventions and produces poor offspring when two dissimilar networks are recombined.

Whitley et al. (1993) compare GENITOR to the Adaptive Heuristic Critic (Anderson, 1989, AHC), which uses the TD method of reinforcement learning. In several different versions of the common pole-balancing benchmark task, GENITOR was found to be comparable to the AHC in both learning rate and generalization. One interesting difference Whitley et al. found was that GENITOR was more consistent than the AHC in solving the pole-balancing problem when the failure signals occurs at wider pole bounds (make the problem much harder). For AHC, the preponderance of failures appears to cause all states to overpredict failure. In contrast, the EA method appears more effective in finding policies that obtain better overall performance, even if success is uncommon. The difference seems to be that the EA tends to ignore those cases where the pole cannot be balanced, and concentrate on successful cases. This serves as another example of the advantages associated with search in policy space, based on overall policy performance, rather than paying too much attention to the value associated with individual states.

## 10.4 SANE

The SANE (Symbiotic, Adaptive Neuro-Evolution) system was designed as a efficient method for building artificial neural networks in RL domains where it is not possible to generate training data for normal supervised learning (Moriarty & Miikkulainen, 1996a, 1998). The SANE system uses an evolutionary algorithm to form the hidden layer connections and weights in a neural network. The neural network forms a direct mapping from sensors to actions and provides effective generalization over the state space. SANE's only method of credit assignment is through the EA, which allows it to apply to many problems where reinforcement is sparse and covers a sequence of decisions. As described previously, SANE uses a distributed representation for policies.

SANE offers two important advantages for reinforcement learning that are normally not present in other implementations of neuro-evolution. First, it maintains diverse populations. Unlike the canonical function optimization EA that converge the population on a single solution, SANE forms solutions in an *unconverged* population. Because several different types of neurons are necessary to build an effective neural network, there is inherent evolutionary pressure to develop neurons that perform different functions and thus maintain several different types of individuals within the population. Diversity allows recombination operators such as crossover to continue to generate new neural structures even in prolonged evolution. This feature helps ensure that the solution space will be explored efficiently throughout the learning process. SANE is therefore more resilient to suboptimal convergence and more adaptive to changes in the domain.

The second feature of SANE is that it explicitly decomposes the search for complete solutions into a search for partial solutions. Instead of searching for complete neural networks all at once, solutions to smaller problems (good neurons) are evolved, which can be combined to form an effective full solution (a neural network). In other words, SANE effectively performs a problem reduction search on the space of neural networks.

SANE has been shown effective in several different large scale problems. In one problem, SANE evolved neural networks to direct or focus a minimax game-tree search (Moriarty

& Miikkulainen, 1994). By selecting which moves should be evaluated from a given game situation, SANE guides the search away from misinformation in the search tree and towards the most effective moves. SANE was tested in a game tree search in Othello using the evaluation function from the former world champion program Bill (Lee & Mahajan, 1990). Tested against a full-width minimax search, SANE significantly improved the play of Bill, while examining only a subset of the board positions.

In a second application, SANE was used to learn obstacle avoidance behaviors in a robot arm (Moriarty & Miikkulainen, 1996b). Most approaches for learning robot arm control learn hand-eye coordination through supervised training methods where examples of correct behavior are explicitly given. Unfortunately in domains with obstacles where the arm must make several intermediate joint rotations before reaching the target, generating training examples is extremely difficult. A reinforcement learning approach, however, does not require examples of correct behavior and can learn the intermediate movements from general reinforcements. SANE was implemented to form neuro-control networks capable of maneuvering the OSCAR-6 robot arm among obstacles to reach random target locations. Given both camera-based visual and infrared sensory input, the neural networks learned to effectively combine both target reaching and obstacle avoidance strategies.

For further related examples of evolutionary methods for learning neural-net control systems for robotics, the reader should see (Cliff, Harvey, & Husbands, 1993; Husbands, Harvey, & Cliff, 1995; Yamauchi & Beer, 1993).

## 11. Summary

This article began by suggesting two distinct approaches to solving reinforcement learning problems; one can search in value function space or one can search in policy space. TD and EARL are examples of these two complementary approaches. Both approaches assume limited knowledge of the underlying system and learn by experimenting with different policies and using reinforcement to alter those policies. Neither approach requires a precise mathematical model of the domain, and both may learn through direct interactions with the operational environment.

Unlike TD methods, EARL methods generally base fitness on the overall performance of a policy. In this sense, EA methods pay less attention to individual decisions than TD methods do. While at first glance, this approach appears to make less efficient use of information, it may in fact provide a robust path toward learning good policies, especially in situations where the sensors are inadequate to observe the true state of the world.

It is not useful to view the path toward practical RL systems as a choice between EA and TD methods. We have tried to highlight some of the strengths of the evolutionary approach, but we have also shown that EARL and TD, while complementary approaches, are by no means mutually exclusive. We have cited examples of successful EARL systems such as SAMUEL and ALECSYS that explicitly incorporate TD elements into their multi-level credit assignment methods. It is likely that many practical applications will depend on these kinds of multi-strategy approaches to machine learning.

We have also listed a number of areas that need further work, particularly on the theoretical side. In RL, it would be highly desirable to have a better tools for predicting the amount of experience needed by a learning agent before reaching a specified level of per-formance. The existing proofs of optimality for both Q-learning and EA are of extremely limited practical use in predicting how well either approach will perform on realistic problems. Preliminary results have shown that the tools of PAC analysis can be applied to both EA an TD methods, but much more effort is needed in this direction.

Many serious challenges remain in scaling up reinforcement learning methods to realistic applications. By pointing out the shared goals and concerns of two complementary approaches, we hope to motivate further collaboration and progress in this field.